%% file: main.tex
\documentclass[runningheads]{llncs}
\usepackage{cite}
\usepackage{amsmath,amssymb,amsfonts}
\usepackage{array}
\usepackage{graphicx}
\usepackage{tabularx}
\usepackage{textcomp}
\usepackage{multirow}
\usepackage{multicol}
\usepackage{diagbox}
\usepackage{hhline}
\usepackage[table]{xcolor}
\usepackage[english]{babel}
\usepackage{adjustbox}
\usepackage[nottoc]{tocbibind}
\usepackage{algorithm}
\usepackage{algpseudocode}
\usepackage[hidelinks]{hyperref}
\usepackage{subcaption}
\usepackage{parskip}
\usepackage{fancyhdr}
\usepackage{blindtext}
\usepackage{listings}
\usepackage{mathtools}
\usepackage{microtype}
\usepackage{calc}     
\usepackage{relsize}  
\usepackage{booktabs}
\usepackage{pdflscape}
\usepackage{rotating}
\usepackage{lscape}
\lstnewenvironment{context}[1][]
{\lstset{
    basicstyle=\normalsize\ttfamily,
    breaklines=true,
    framexleftmargin=0mm,
    frame=none,
    backgroundcolor=\color{gray!10},
    #1
  }}
{}

\let\llncssubparagraph\subparagraph
\let\subparagraph\paragraph
\usepackage[compact]{titlesec}
\let\subparagraph\llncssubparagraph
\titlespacing*{\subsubsection}{0pt}{2.0ex plus 0.5ex minus .2ex}{0.8ex plus .2ex}

\begin{document}

\title{Generative AI-Based Virtual Assistant Using Retrieval-Augmented Generation: \\ \normalsize An evaluation study for bachelor projects}

\author{Dumitru Verşebeniuc\inst{1}\orcidID{0009-0004-4660-9636} \and
Martijn Elands\inst{1}\orcidID{0009-0004-1296-5413} \and
Sara Falahatkar\inst{1} \and
Chiara Magrone\inst{1} \and
Mohammad Falah\inst{1} \and
Martijn Boussé\inst{1}\orcidID{0000-0003-2090-0682} \and
Aki Härmä\inst{1}\orcidID{0000-0002-2966-3305}
}

\titlerunning{Generative AI-Based Virtual Assistant using Retrieval-Augmented Generation: \\ \normalsize An evaluation study for bachelor projects}
\authorrunning{D. Verşebeniuc et al.}
%

\institute{Department of Advanced Computing Sciences, Maastricht University, Maastricht 6200 MD, The Netherlands 
\\ 
\email {d.versebeniuc, m.elands, c.magrone, s.falahatkar, m.falah@student.maastrichtuniversity.nl} \\
\email {m.bousse, aki.harma@maastrichtuniversity.nl}
}

\maketitle

\begin{abstract}
Large Language Models have been increasingly employed in the creation of Virtual Assistants due to their ability to generate human-like text and handle complex inquiries. While these models hold great promise, challenges such as hallucinations, missing information, and the difficulty of providing accurate and context-specific responses persist, particularly when applied to highly specialized content domains. In this paper, we focus on addressing these challenges by developing a virtual assistant designed to support students at Maastricht University in navigating project-specific regulations. We propose a virtual assistant based on a Retrieval-Augmented Generation system that enhances the accuracy and reliability of responses by integrating up-to-date, domain-specific knowledge. Through a robust evaluation framework and real-life testing, we demonstrate that our virtual assistant can effectively meet the needs of students while addressing the inherent challenges of applying Large Language Models to a specialized educational context. This work contributes to the ongoing discourse on improving LLM-based systems for specific applications and highlights areas for further research.\\
\end{abstract}


\section{Introduction} \label{sec:introduction}

In this paper, we propose a virtual assistant (VA) that combines recent techniques from Natural Language Processing (NLP) and Retrieval-Augmented Generation (RAG). The VA contains a multi-query and self-reflection mechanism and was tested with students from the Department of Advanced Computing Sciences (DACS) of Maastricht University as its goal was to address common challenges faced by students when seeking information about academic rules and regulations related to bachelor projects. For example, if a student cannot attend meeting X due to reason Y, the student needs to review the rules and regulations for meeting X to determine if reason Y is acceptable. This process can be time-consuming/unsuccessful for the student and may result in contacting the project coordinator. The VA aims to provide accurate and contextually relevant responses to student inquiries, improving the student experience and addressing information overload challenges. Even though we limited ourselves to the application of a VA within bachelor projects, our approach can be generalized easily to various academic settings.

Pre-trained language models have revolutionized NLP, enabling efficient information storage and retrieval with impressive accuracy \cite{csedu20}. However, Large Language Models (LLMs) often struggle with precise and contextually sensitive knowledge manipulation \cite{Lewis}. In this paper, we address this limitation by using the RAG framework, integrating retrieval mechanisms that access up-to-date and domain-specific information from external sources, and enhancing the generative capabilities of LLMs \cite{RAG_survey_july2024}.

As students might struggle with phrasing their questions precisely, we implemented a multi-query mechanism, thereby improving the retrieval process and ensuring the most relevant answers are provided. Moreover, generative LLMs can hallucinate while generating responses. In such cases, we have a self-reflection system that evaluates the response and attempts to correct it if the information is incorrect or if the answer does not align with the question. The combination of these features creates a novel approach that underscores our VA's potential to enhance the field of academic assistance tools. Moreover, rather than relying solely on predefined testers, we conducted trials with actual bachelor students, gathering their feedback to refine the system's performance in real-world scenarios. (\href{https://youtu.be/MMNZSac48_k?si=sAMlNoSbOYrQFXtw}{A demonstration of the VA's functionality can be found here.})

Building upon foundational work in RAG and knowledge graph-based systems like the GRAPE model \cite{grape} and Amazon's VA for financial reconciliation \cite{Amazon}, this VA aspires to deliver contextual information to the student query in an academic environment. Our approach is aligned with the broader trends in RAG, as surveyed extensively by Wang et al. \cite{wu2024retrievalaugmentedgenerationnaturallanguage}, who provide a detailed overview of how RAG techniques are enhancing various NLP tasks. However, our VA distinguishes itself by specifically addressing the challenges faced by students in an academic setting, by incorporating novel strategies to better interpret and respond to student inquiries.

For this showcase, the goal of the VA is to alleviate increased workload of staff due to the addition of a new bachelor program in the Department of Advanced Computing Sciences (DACS) at Maastricht University. In particular, this caused a doubling of the number of project groups (from 36 to 72) with 6-7 students per group, putting immense pressure on project coordinators and tutors in managing student inquiries about project organization, rules and regulations, and examination details. For example, coordinators were contacted multiple times with similar, yet subtlety different, questions such as "What are the criteria for X in case of Y?" or "What happens if I miss X, but did do Y". By focusing on project-specific information and leveraging an LLM, the VA can assist students by providing immediate answers to these common questions, reducing the burden on staff while still offering a personalized experience.

Project Preview: Bachelor projects at Maastricht University involve students working in small groups of about six to seven persons, guided by a fixed tutor, on a project divided into three subtasks across three periods. Students work part-time and full-time at different stages, concluding each period with presentations and submissions, culminating in a final report and product examination.

We conducted a pre-survey to assess student needs for a system like this before developing the VA. From 386 first-year bachelor students, 27 participants responded, with 75\% indicating they had previously consulted project coordinators about project organization and rules. Participants showed a strong preference for such a system, with an average rating of 4.2 and a median of 5 on a 5-point Likert scale.

We aim to answer the following questions regarding the VA's performance:
\begin{enumerate}
    \item How accurately can the VA retrieve relevant content to the student's queries?
    \item How precisely can the VA generate a relevant response to the student's queries given the retrieved content?
    \item What is the fallback mechanism employed by the VA when unable to retrieve a suitable answer to a student's question?
    \item What is the average response time of the VA in providing answers to student queries while ensuring the quality and comprehensiveness of the response content?
\end{enumerate}

This paper is organized as follows. In Section~\ref{sec:methods}, we explain our methodology, including retrieval and generation pipeline as well as self-reflection. In Section~\ref{sec:metrics} and~\ref{sec:study_design}, we outline the evaluation process and the experiments. Results are reported in Section~\ref{sec:results}. Discussion and conclusion can be found in Section~\ref{sec:discussion} and Section~\ref{sec:conclusion}.

\section{Methods} \label{sec:methods}
\begin{figure*}[!ht]
  \centering
  \includegraphics[width=\linewidth]{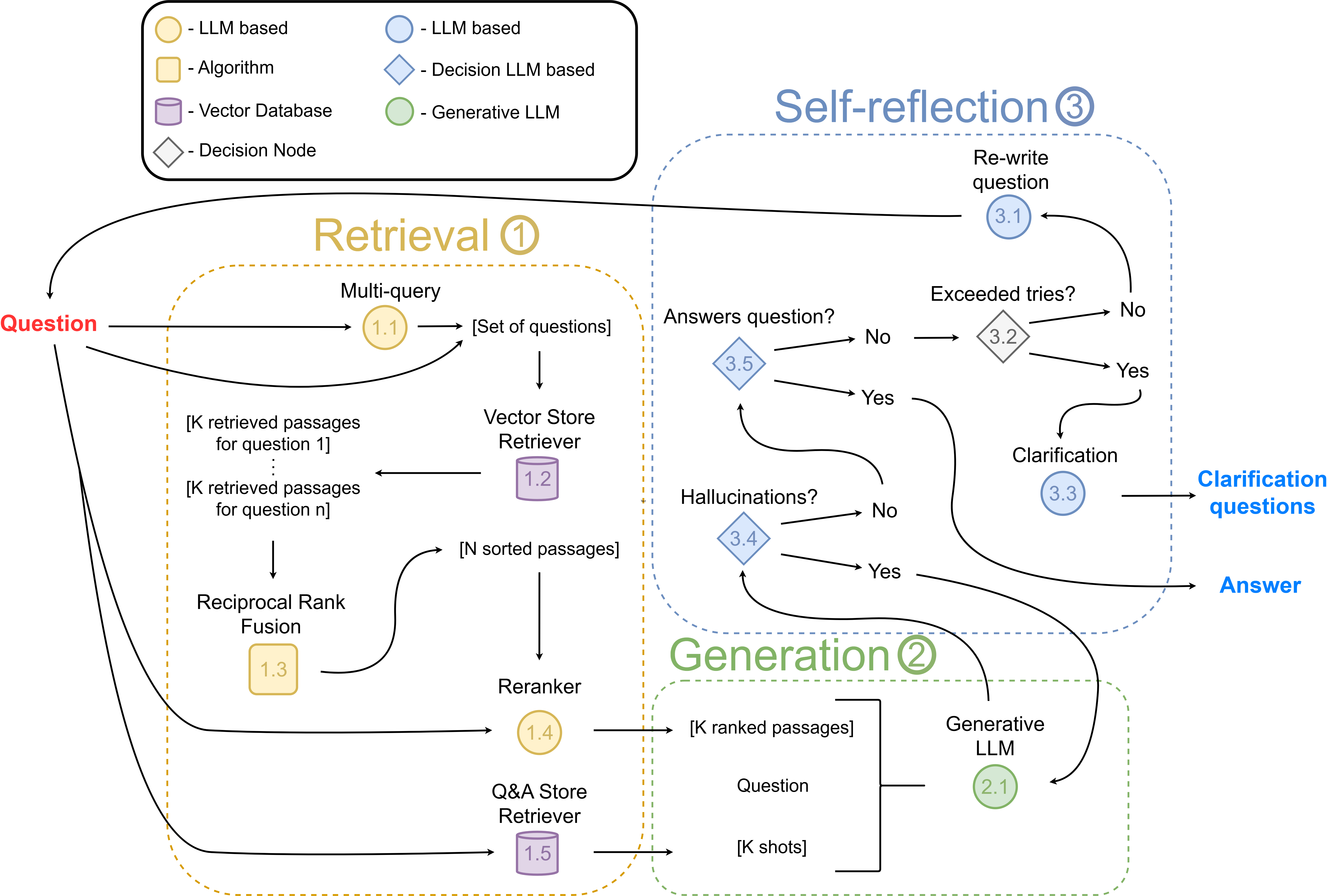}
  \caption{The VA architecture consists of retrieval, generation, and self-reflection parts. Retrieval \raisebox{1pt}{\textcircled{\raisebox{-.9pt} {1}}} collects the relevant information to the user question. Generation \raisebox{1pt}{\textcircled{\raisebox{-.9pt} {2}}} processes the retrieval information and structures it to generate the response. Self-reflection \raisebox{1pt}{\textcircled{\raisebox{-.9pt} {3}}} is the fallback mechanism that ensures the correctness of the response and understanding of the user question.}
  \label{fig:architecture}
\end{figure*}

The following section details the architecture, processes, and methodologies employed in the VA pipeline (see Figure \ref{fig:architecture}). The VA pipeline receives a user question as input to the retrieval pipeline Figure \ref{fig:architecture}.1. The retrieval pipeline collects all necessary information such as relevant document chunks and similar Question and Answer (Q\&A) examples to the user question, and then sends all collected information to the generation pipeline Figure \ref{fig:architecture}.2. The generation pipeline organizes the information from the retrieval phase and generates the response with a set of instructions, which it then sends to the self-reflection part Figure \ref{fig:architecture}.3.

The self-reflection part is our fallback mechanism, which evaluates the generated response for hallucinations and relevance. If one part of self-reflection fails, it attempts to correct itself or asks for clarifications from the user. The detailed mechanisms of the VA pipeline are discussed later in this paper, with numerical references as illustrated in Figure \ref{fig:architecture}.

\textit{Retrieval-Augmented Generation:}
RAG is a technique that combines the strengths of both retrieval-based and generation-based methods. This hybrid method involves retrieving relevant documents from a large corpus and using this information to generate more accurate and contextually enriched responses \cite{RAG}. The integration of RAG allows our VA to access up-to-date information, thus overcoming the temporal limitations of static pre-trained models.

\textit{Manual Parsing:} Since we are dealing with a not large amount of data, we consider manual parsing, where we split documents into relevant chunks, which helps avoid including noise during the retrieval process.

\subsection{Retrieval Pipeline \raisebox{1pt}{\textcircled{\raisebox{-.9pt} {1}}}} \label{sec:retriever}
The retrieval pipeline is a fundamental component of our VA system, designed to efficiently retrieve relevant information from a corpus of documents.
\subsubsection{Multi-Query Retrieval \raisebox{1pt}{\textcircled{\raisebox{0.2pt} {\tiny 1.1}}}} Multi-query retrieval enhances the retrieval process \cite{iter} by generating multiple versions of a user query, capturing different perspectives, and ensuring a broader range of relevant documents. This approach leverages LLMs to create alternative queries, improving the accuracy and comprehensiveness of the VA's responses by considering diverse and contextually relevant documents.
\subsubsection{Vector Database \raisebox{1pt}{\textcircled{{\tiny 1.2}}}}
The vector database is the whole basis of our retrieval system, enabling efficient storage and retrieval of document embeddings \cite{Han2023-xp}. The embedding model transforms documents and queries into dense vector representations, ensuring quick and accurate retrieval of relevant documents based on similarity metrics such as cosine similarity, maximal marginal relevance (MMR), etc.
\subsubsection{Embedding Techniques}
Embedding techniques are crucial for converting text into numerical vectors that models can understand and manipulate. We can use advanced embedding models from Google \cite{googleTextEmbeddings}, OpenAI \cite{openaiTextEmbeddings}, Mistral \cite{mistralEmbeddingsMistral}, or BGE \cite{chen2024bge} to generate high-quality numeric text representation. These embeddings capture semantic meanings and relationships within the text \cite{Patil2023-fv}, which is essential for both the retrieval and generation phases.
\subsubsection{Reciprocal Rank Fusion \raisebox{1pt}{\textcircled{\raisebox{0.5pt} {\tiny 1.3}}}}
Reciprocal Rank Fusion (RRF) is a method used in RAG to combine and rerank documents from multiple retrieval queries. By assigning a score to each document based on its ranking across various result lists, RRF effectively consolidates different retrieval outputs. The formula for RRF is given by:
\[
\text{score}_{\text{RRF}}(D) = \sum_{i=1}^{k} \frac{1}{k + \text{rank}_i(D)}
\]
where \( \text{rank}_i(D) \) is the rank of document \( D \) in the \( i \)-th result list, and \( k \) is a small constant, often set to 60. This method enhances retrieval accuracy by emphasizing higher-ranked documents while maintaining robustness against discrepancies in individual list rankings.
\subsubsection{Reranker \raisebox{1pt}{\textcircled{\raisebox{0.5pt} {\tiny 1.4}}}} The reranker model is used to obtain a better relevance score between a question and a document. There are two different types of reranked models:

\begin{itemize}
    \item\textit{Cross-encoder model} uses both the question and the document as inputs and directly outputs a similarity score, rather than generating embeddings. \cite{chen2024bge}

    \item\textit{LLM-Embedder}, unlike embeddings, which primarily assess semantic similarity between a document and a query, the LLM-Embedder can provide precise scores for how well a document answers a given query using fine-tuned generative LLMs. \cite{zhang2023retrieve}

\end{itemize}

\subsubsection{Few-shot retriever \raisebox{1pt}{\textcircled{\raisebox{0.3pt} {\tiny 1.5}}}}
Since we have already received a Q\&A dataset\footnote{The Q\&A dataset consists of a set of questions posed by students and answers provided by the project coordinator. The questions are typically inquiries about project-related matters, such as assessment, attendance, or deliverables, while the answers provide feedback or clarification on those inquiries.} from the bachelor's project coordinator, storing them in the Vector Store allows us to retrieve similar questions and example truth answers to provide few-shot examples to our generative LLM, thereby giving example answers to the question.

\subsection{Generation Pipeline \raisebox{.5pt}{\textcircled{\raisebox{-.9pt} {2}}}}
In the generation process of a VA, using low-temperature 0.2 and structured XML prompts (\href{https://github.com/DikaVer/maastricht_university_generative_virtual_assistant}{shared on our GitHub}) helps to improve answer accuracy and factual grounding. Low temperatures reduce randomness, leading to more conservative and reliable responses, while structured XML prompts facilitate better input understanding and contextual relevance by clearly defining different elements of a prompt. This combination ensures that the LLM \raisebox{1pt}{\textcircled{\raisebox{0.3pt} {\tiny 2.1}}} generates responses that are consistent and closely aligned with the provided facts. Moreover, this technique optimizes the VA's functionality in handling detailed and complex queries \cite{promptEng}.

\subsection{Self-Reflection \raisebox{.5pt}{\textcircled{\raisebox{-.9pt} {3}}}}
The self-reflection process is about assessing and refining the response generated by the VA to obtain accuracy and relevance \cite{Jeong2024-pk}. Here's how it functions:

\begin{itemize}
    \item \textit{Re-write Question} \raisebox{1pt}{\textcircled{\raisebox{0.3pt} {\tiny 3.1}}}: If the generated answer is not satisfactory, or if it does not directly address the user's query, the process includes a mechanism to rewrite the query. This might involve rephrasing, correcting, or breaking down the question into more manageable parts to match the available data better. By rephrasing the question, the VA can better match the query with relevant documents in the database, leading to more accurate responses.
    
    \item \textit{Exceeded Tries} \raisebox{1pt}{\textcircled{\raisebox{0.2pt} {\tiny 3.2}}}: This checks whether the process of rewriting the question has exceeded a set number of tries. If so, it indicates that the system is struggling to understand the query or to find relevant information, and a different approach might be needed such as clarification questions. 
    
    \item \textit{Clarification Questions} \raisebox{1pt}{\textcircled{\raisebox{0.3pt} {\tiny 3.3}}}: If necessary, the system can pose clarification questions to the user. This step is especially important when the query is ambiguous or lacks specific details needed for an accurate response. By engaging the user in a dialogue, the VA can gather additional information to better address the user's needs.
    
    \item \textit{Hallucinations Check} \raisebox{1pt}{\textcircled{\raisebox{0.3pt}{\tiny 3.4}}}: This step involves checking whether the generated responses include hallucinated information, which is a common issue with generative models. This is achieved using a generative LLM as a decision-maker to evaluate the generated answer against the retrieved documents.

    The hallucination detection mechanism works by comparing the response to the set of facts obtained during the retrieval phase. The generative LLM uses a specific prompt to determine whether the answer is grounded in the provided facts.
    
    If the LLM determines that the answer contains information not supported by the facts, it requests the Generative LLM \raisebox{1pt}{\textcircled{\raisebox{0.3pt}{\tiny 2.1}}} to regenerate the response until the Hallucination Checker \raisebox{1pt}{\textcircled{\raisebox{0.3pt}{\tiny 3.4}}} produces a satisfactory result.

    \item \textit{Answer Check} \raisebox{1pt}{\textcircled{\raisebox{0.3pt} {\tiny 3.5}}}: This involves verifying whether the generated response adequately addresses 
    the user's question. The system employs a generative LLM as a decision-maker to determine if the answer resolves the query. If the answer is found lacking, the system may trigger the query rewriting mechanism \raisebox{1pt}{\textcircled{\raisebox{0.3pt} {\tiny 3.1}}} or ask for clarification \raisebox{1pt}{\textcircled{\raisebox{0.3pt} {\tiny 3.3}}} from the user.
\end{itemize}

\input{sections/metrics} 

\section{Experiments} \label{sec:study_design}
A series of tests with DACS bachelor students were performed to ensure that the system was user-friendly and to identify any other improvements. Furthermore, we also wanted to gather quantitative measurements about the system's performance. These tests were conducted on June 20 and June 21, 2024.

The students who participated were divided into two groups: A and B. Each group received eight scenarios accompanied by a multiple-choice question to assess their knowledge of the rules and regulations and examination details. An ``I don't know'' option was added to limit guessing behavior. For the first four scenarios, participants in Group A could not use the VA, while participants in Group B were encouraged to use the VA. For the last four scenarios, the roles were reversed, making the tests complementary. We ensured no leakage occurred. 

At the start of the test, participants received a list of basic background questions. During the test, the first four scenarios also featured a question asking if the participant would prefer to ask the VA or the coordinator first. The last four scenarios are accompanied by a question if the VA helped answer the scenario on a Likert scale where 1 was not helpful at all and 5 was very helpful. At the end of the test, they were asked some general questions, for example, about the response time.

For example, scenario 3 is: ``Luca carefully planned his day to arrive on time for the final product and report examination. However, the bus he took left the stop earlier than scheduled, causing him to miss the exam. Now he is worried about his project grades. What are the consequences for his project grades?"

Answers:
\begin{itemize}
    \item \textbf{He will receive a NG for the project}
    \item He will receive the same grade as everyone else in the team
    \item He will receive a lowered individual grade for the project
\end{itemize}

The full survey is shared online on our \href{https://github.com/DikaVer/maastricht_university_generative_virtual_assistant}{GitHub page}.

\section{Results} \label{sec:results}

\subsection{Automated Retrieval Augmented Generation Assessments}
\autoref{tab:metric_results} shows the summary of the metrics presented in section \ref{sec:metrics} for the system with different generative LLMs. While Gemini 1.0 Pro slightly outperforms GPT-3.5 in context precision, GPT-3.5 excels in other key areas like context recall, answer relevancy, and faithfulness. This suggests that GPT-3.5 generates more relevant and accurate content, making it better suited for our VA.

\begin{table}[htbp]
\centering
\caption{Evaluation metrics for different generative LLMs}
\begin{tabular}{>{\raggedright}p{5cm}p{2cm}p{2cm}}\toprule
Metric               & GPT3.5 & Gemini1.0Pro \\ \hline
Context precision    & 88\%          & \textbf{89\%}          \\
Context recall       & \textbf{42\%}          & 41\%          \\
Answer relevancy     & \textbf{57\%}          & 37\%          \\
Faithfulness         & \textbf{43\%}          & 32\%          \\
Customized precision & \textbf{77\%}          & \textbf{77\%}          \\ \hline
\end{tabular}
\label{tab:metric_results}
\end{table}

\subsection{Survey Results}
In total, 64 participants completed the survey: 34 in group A and 30 in group B. As described in Section \ref{sec:study_design}, participants were asked about their enrolled program, year of study, and whether they had ever read the rules and regulations. We found that 55\% of the participants had read the rules and regulations.

Participants were also asked how they would respond to the first four presented scenarios: whether they would contact the project coordinator or a VA for assistance. On average, 71\% of participants preferred contacting the project coordinator before the VA. This preference was consistent across all scenarios, even without prior use of the VA.

\subsubsection{Students' Knowledge of Project Rules:}
The survey revealed varied levels of understanding among participants regarding the rules for skipping project meetings. However, based on responses, both groups demonstrated similar prior knowledge. This suggests that misunderstandings about the rules and regulations were widespread and not confined to any particular group. The correct answer is that one meeting can be skipped in phases 1 and 2 combined, one meeting in phase 3 without consequence, two meetings result in a lower project grade, and three meetings result in an NG. Out of the respondents, only 7 participants had the correct understanding of these rules.

Additionally, 28 participants believed that one meeting could be skipped per phase, which shows a partial understanding of the rules but lacks specificity regarding phase combinations and consequences. Another 11 participants thought that one meeting could be skipped without consequences, with the second and third meetings leading to a grade reduction and NG respectively, which again is partially correct but not entirely accurate.

There was also a group of 5 participants who admitted to not knowing the rules, indicating a clear gap in knowledge. Furthermore, some participants provided detailed answers that did not fully align with the correct rules, showing a mixture of partial knowledge and misconceptions about the consequences of missing project meetings. This data suggests that while most participants have a general idea about the meeting policies, there is significant room for improvement in ensuring that all participants have a precise and thorough understanding of the rules.

\subsubsection{Scenario Results:} In \autoref{tab:va_performance}, we show the results for all scenarios with and without VA assistance. The data indicates an overall improvement in performance when participants used the VA. Performance improvement when using the VA can be seen based on the data collected from participants' responses to the scenarios.

\begin{table}[h]
\centering
\caption{Impact of VA on Correct Answers and ``I Don't Know" Responses showing improvements in accuracy and reducing uncertainty in student responses across most scenarios.}
\label{tab:va_performance}
\begin{tabular}{>{\raggedright}p{2cm}*{4}{p{2.5cm}}}\toprule
\multirow{2}{*}{\textbf{Scenarios}}  &  \multicolumn{2}{c}{Correct Answers}  & \multicolumn{2}{c} {``I Don't Know"} 
\\\cmidrule(ll){2-3}\cmidrule(ll){4-5}
& Without VA & With VA  & Without VA & With VA \\\midrule
1 & 32.4\% & 46.7\% & 29.4\% & 16.7\% \\
2 & 17.6\% & 26.7\% &  32.4\% &  6.7\% \\
3 & 32.4\% & 56.7\% & 50.0\% & 16.7\% \\
4 &  2.9\% & 26.7\% & 41.4\% & 26.7\% \\ 
5 & 53.3\% & 35.3\% &  30.0\% &  2.9\% \\
6 & 50.0\% & 73.5\% &  30.0\% &  5.9\% \\
7 & 63.3\% & 50.0\% &  3.3\% &  2.9\% \\
8 & 36.7\% & 32.4\% & 30.0\% & 32.4\% \\
\bottomrule
\end{tabular}
\end{table}

The results indicate that, in general, the performance of participants improved when using the VA, as evidenced by the higher percentages of correct answers in most scenarios. For instance, in Scenario 1, the percentage of correct answers increased from 32.4\% without the VA to 46.7\% with the VA. Similarly, in Scenario 6, the correct answers jumped from 50\% to 73.5\%. These improvements suggest that the VA can assist students in answering organizational questions related to rules and regulations.
However, there are notable exceptions, such as Scenarios 5 and 7, where the performance did not improve as expected. In Scenario 5, the percentage of correct answers decreased from 53.3\% without the VA to 35.3\% with the VA. Similarly, in Scenario 7, the correct answers decreased from 63.3\% to 50\%. The reasons why that might have happened are explained in \ref{subsec:Refinements}. 

Additionally, the ``I don't know" responses generally decreased with the use of the VA, indicating that the assistant helped reduce uncertainty among participants. For example, in Scenario 2, the ``I don't know" responses dropped from 32.4\% without the VA to 6.7\% with the VA. This further supports the utility of the VA in providing clearer guidance and information to the students.

When scenarios 5 and 7 are left out-of-scope for reasons mentioned in section \ref{subsec:Refinements}, we can state that we have significantly improved the scenarios with a 95\% confidence interval for the correct answers and the reduction of ``I don't know" submissions over the usage without a VA.

\subsubsection{Scenario Feedback:}
\autoref{tab:score_distribution} presents the distribution of helpfulness scores received by the VA for various scenarios. Participants rated the VA on a Likert scale ranging from 1 (not helpful) to 5 (extremely helpful). This table allows us to analyze how the VA's perceived helpfulness varies across different tasks it was asked to perform.

\begin{table}[htbp]
\centering
\caption{Distribution of Helpfulness Scores of VA for each Scenario}
\label{tab:score_distribution}
\renewcommand{\arraystretch}{1.2} 
\begin{tabular}{>{\raggedright}p{2cm}*{5}{p{2cm}}} 
\toprule
\multirow{2}{*}{\textbf{Scenarios}} &  \multicolumn{5}{c}{\textbf{On a scale of 1 to 5, how helpful was the VA?}}  \\
\cmidrule(lllll){2-6}
& 1 & 2  & 3 & 4 & 5 \\
\midrule
1 & 6.67\% & 16.67\% & 30\% & 20\% & 26.67\% \\
2 & 10\% & 16.67\% &  33.33\% &  26.67\% & 13.33\% \\
3 & 10\% & 16.67\% & 13.33\% & 26.67\% & 33.33\% \\
4 & 6.67\% & 16.67\% & 40\% & 30\% & 6.67\% \\ 
5 & 11.76\% & 5.88\% &  8.82\% &  29.41\% & 44.12\% \\
6 & 0.00\% & 8.82\% &  2.94\% &  47.06\% & 41.18\% \\
7 & 0.00\% & 8.82\% &  17.65\% &  35.29\% & 38.24\% \\
8 & 2.94\% & 20.59\% & 20.59\% & 32.35\% & 26.67\% \\
\bottomrule
\end{tabular}
\end{table}

Scenarios 5 and 6, show a significant portion of participants (over 40\%) rating the VA as extremely helpful (score 5). This suggests the VA effectively assisted users in those specific situations. In contrast, scenarios with a wider range of scores, such as Scenario 1 and 8, indicate a more diverse range of user experiences. This implies a mixed perception of the VA's usefulness in those Scenarios, which might be caused by a difference of opinion between students and the rules and regulations.

\subsubsection{Overall Feedback Experience:}
\begin{figure*}[htbp]
  \centering
  \includegraphics[width=0.8\linewidth]{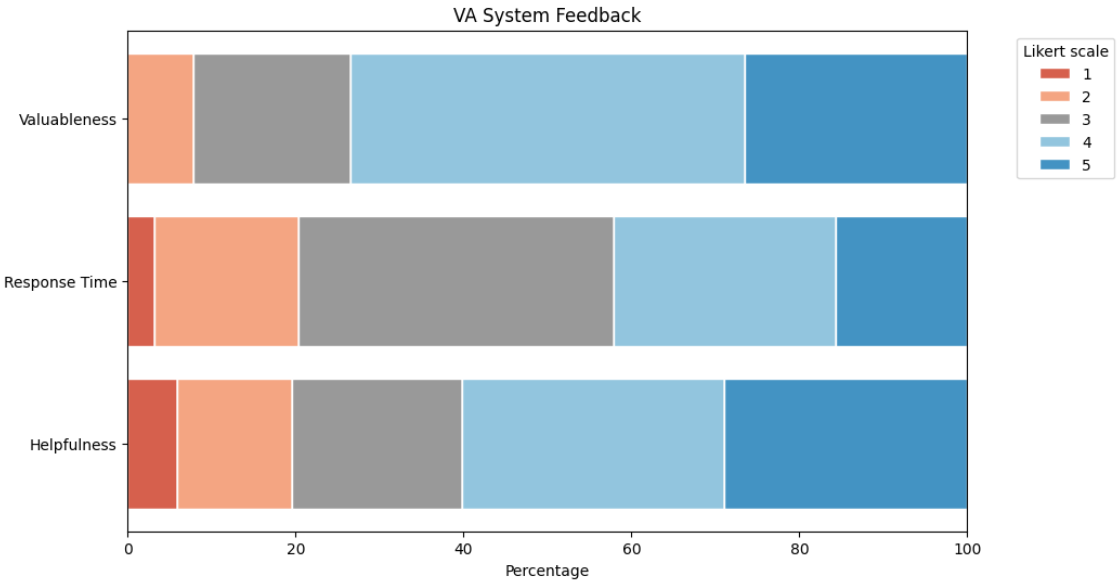}
\caption{Overall student satisfaction with the VA system showing a generally positive reception, based on students' rating}
  \label{fig:overall_experience}
\end{figure*}

Figure \ref{fig:overall_experience} shows the overall satisfaction of students with the VA system. Each bar represents a summary of the Likert scale responses for three different questions:

\begin{itemize}
    \item ``Do you think that we developed a valuable VA? By valuable, we mean that it can reduce staff members' workload and increase response time for students." This is also on a Likert scale where 1 is ``Not valuable at all" and 5 is ``Extremely valuable".
    \item ``How was the response time?" This is on a Likert scale where 1 is ``Worse than expected" and 5 is ``Better than expected".
    \item In the third case, we summarized the responses to the question posed to students after each scenario, ``Was the VA helpful in answering this question?" This was also rated on a Likert scale.
\end{itemize}

The chart shows the distribution of responses for each category, indicating a generally positive reception to the VA system. For response time, students predominantly rated it as ``Neutral", ``Satisfied", and ``Very satisfied", reflecting a positive view of the VA's promptness. The perceived value of the VA was also highly rated, with most students finding it valuable for reducing staff workload and improving response times. In evaluating the VA's helpfulness in answering specific questions, the majority of responses were ``Satisfied" and ``Very satisfied", demonstrating the VA's effectiveness across various scenarios. Overall, the feedback highlights a positive reception, appreciating both the response time and added value of the VA, while also suggesting areas for further improvement.

\begin{table}[h]
\caption{The table categorizes the most asked topics related to group projects, revealing the frequency of specific concerns among students. It identifies the most discussed issues, focusing on areas that may require more attention.}
\begin{tabular}{>{\raggedright}p{1.5cm}p{9cm}p{1.5cm}}\toprule
\textbf{Topic} & \textbf{Representative Words} & \textbf{Counts}\\
\midrule 
0 & [Missed pre-examination and final presentation consequences] & 97 \\
1 & [Dealing with Inactive Group Members in a Project] & 23 \\
2 & [help, context, new update] &  12 \\
3 & [Force Majeure and Attendance Issues] & 9 \\
4 & [Factors affecting individual grades in projects] & 8 \\
5 & [Communication with Tutors and Examiners] & 4 \\
6 & [What happens when the whole group misses something] & 4 \\
7 & [Consequences of a given case] & 4\\
8 & [Phase 3 Consequences] & 4 \\\bottomrule
\end{tabular}
\label{tab:topics}

\end{table}

\subsubsection{Response-Time Analysis:}
The analysis of response times indicated that the VA achieved an average response time of 10.045 seconds across all queries, with a standard deviation of 2.39 seconds. While this performance was satisfactory for the purposes of initial testing, there is room for improvement to ensure more consistent and reliable response times in real-life deployment scenarios. For instance, during periods of high demand, such as when multiple queries are submitted simultaneously, the VA exhibited increased response times, suggesting the need for enhanced handling of concurrent requests to maintain efficiency.

\subsection{Refinements}\label{subsec:Refinements}
Student testing revealed areas for improvement in the VA's performance. For example, Scenario 5 highlighted the need for more comprehensive training data, as the VA struggled to distinguish between a project meeting and an exam due to missing information in the reference document. Similarly, Scenario 7 showed the limitations of the VA in handling unforeseen scenarios not covered by the rules. These findings suggest a need to refine the training data and potentially develop mechanisms for the VA to handle situations outside its current knowledge base.
Additionally, the feedback received from students provides valuable insights for improving the system to better align with their needs. This feedback will be shared in detail on our \href{https://github.com/DikaVer/maastricht_university_generative_virtual_assistant}{GitHub page}.

\section{Discussion} \label{sec:discussion}


    
    
    
    

While our VA shows promising results, we identified several limitations during its development, testing, and design phases.

Firstly, we had to disable the copy-pasting functionality on the client side after the second day of testing. This decision was driven by our observation that students were not following the instructions for paraphrasing or formulating their own questions about the given scenarios. 

Another challenge we faced was determining the appropriate evaluation metrics to use in this setting. In our study, we decided to utilize the RAGAS framework, along with a customized precision metric, to assess the system's performance. However, the field is rapidly evolving, with numerous metrics available, which makes it difficult to choose the most suitable ones.

Additionally, we found that traditional Large Language Models (LLMs) struggle with processing long input texts, as noted in previous research \cite{llm_long_context}. This limitation can result in a loss of detail, which is particularly problematic when evaluating the system's responses. The coordinator's general answers, which tend to be shorter, are often less detailed than the generated responses, leading to potential inconsistencies in evaluation.

During testing, we also observed that the system occasionally misidentifies courses as skill classes. This issue suggests that the VA may benefit from being provided with a predefined list of skill class names for verification purposes. Such an enhancement could reduce the frequency of these misidentifications and improve the system's overall accuracy.

Finally, we encountered a limitation related to our inability to access personal information, such as students' years or their respective coordinators. This lack of information reduces our ability to fully tailor the system to individual student cases. However, this challenge could potentially be overcome by integrating the VA with Maastricht University's Learning Management System or by adding a feature to the User Interface that allows users to input their contact information directly.

\section{Conclusions} \label{sec:conclusion}
The development of a VA for DACS students has shown potential in alleviating the workload of staff and providing a significant impact on fast and accurate information to students. By leveraging advanced NLP techniques and integrating RAG systems, the VA can effectively support students with queries related to rules, regulations, and examination details. The retrieval pipeline, incorporating various embedding models and advanced retrieval techniques, ensures the accuracy and relevance of the information provided.

Our evaluation metrics showed that we can accurately retrieve documents with a context precision of 88\% and a context recall of 42\%. Generating relevant answers based on these documents is done at around 57\% (answer relevancy) and faithfulness of 43\%. All of this is done by the system in around 10 seconds which the students seem to be satisfied with.

By integrating self-reflection, the system can evaluate its own outputs and decision-making processes, allowing it to identify inefficiencies or errors and adjust its methods accordingly. This process is essential for maintaining the reliability of the system; if uncertainties or ambiguities arise, the system can request clarifications, thereby preventing potential errors and refining its responses.

Testing with students has provided valuable insights into the practical application of the assistant. This has led to improvements in the system and indicators for further possible enhancement. Despite its limitations, the VA represents a significant step forward in educational technology, enhancing the student experience and reducing the administrative burden on academic staff. Future work will focus on addressing the identified limitations and exploring additional functionalities to further improve the system's performance and user satisfaction.

Future VA iterations should integrate feedback from the testing participants to align more closely with their expectations, including adjustments like incorporating FAQs and relevant contact information. An example of this is a force majeure template. Additionally, the approach to verify model hallucinations using semantic entropy \cite{Farquhar2024} and employing techniques like Reverse HyDE \cite{Gao2022-fk} and contrastive learning \cite{zhang2023retrieve} will refine the retrieval accuracy and relevance of information provided by the VA.

\bibliographystyle{unsrt}
\bibliography{references}


\end{document}

%% file: sections/metrics.tex
\section{Evaluation} \label{sec:metrics}
We evaluated the VA's performance in the first-year second-semester bachelor projects. In particular, we assessed its ability to retrieve and generate accurate and relevant responses to student queries, as well as its overall effectiveness and user satisfaction.
The vector search evaluation involves assessing the effectiveness of different embedding models and retrieval configurations. The model's performance was evaluated based on its ability to retrieve relevant documents for a given query. The evaluation parameters included search type (similarity, MMR, or similarity score threshold) and various keyword arguments to fine-tune the retrieval process.

In addition to the experiments, we used an offline dataset to evaluate both parts of the pipeline. For this, we use metrics from the RAGAS evaluation framework \cite{es2023ragas} and a custom precision\ref{customized precision} metric.
The evaluation process was automated using a script that iterated through the dataset of question-answer pairs, retrieved relevant documents, and generated responses to calculate the metrics. The next two subsections outline the evaluation metrics.

\subsubsection{Retrieval evaluation}
To measure the performance of the retrieval part, we processed each question to retrieve the most relevant documents. The metrics used for this part are Context Precision and Context Recall.

Context precision is a measure used to see how well a system ranks important pieces of information. Ideally, the most relevant pieces of information (called chunks) should be in a high rank \cite{ragas-metrics}: 
\begin{equation}
    \text{Context Precision@K} = \frac{\sum_{k=1}^{K} (\text{Precision@k} \times v_k)}{\text{total number of relevant items in top } K \text{ results}}
\end{equation}
where
\(\text{Precision@k} = \frac{\text{true positives@k}}{\text{true positives@k} + \text{false positives@k}}\)

Context Recall measures the extent to which retrieved context aligns with the ground truth (annotated answer) \cite{ragas-metrics}:

\begin{equation}
    \text{Context Recall} = \frac{\lvert \text{GT sentences that can be attributed to context} \rvert}{\lvert \text{Number of GT sentences} \rvert}
\end{equation}
where Ground Truth (GT) refers to an annotated or expected correct answer.

Furthermore, we also used a customized precision metric (see Equation \ref{eq: customized precision}) to evaluate the proportion of correctly retrieved documents by the system with the manually labeled correct documents.\label{customized precision}

\begin{equation}\label{eq: customized precision}
    \text{Custom precision} =
    \frac{\lvert \text{correctly retrieved documents} \rvert}{\lvert \text{manually considered relevant documents} \rvert}
\end{equation}
where relevant documents are manually incorporated into the dataset.

\subsubsection{Generation evaluation}
Once the relevant documents were retrieved, the next step involved generating responses using the retrieved content(s). The evaluation metrics for this are Answer Relevancy and Faithfulness. These metrics guarantee the relevancy, accuracy, and fidelity of the generated responses. The generation process was repeated for different generative LLMs \cite{chen2023benchmarking}.

Answer Relevancy assesses how pertinent the generated answer is to the given prompt \cite{ragas-metrics}. It uses artificial questions based on the generated answer. Hence, it uses the mean cosine similarity between the original question and a number of artificial questions.\

\vspace{-7mm}
\begin{equation}
    \text{Answer Relevancy} = \frac{1}{N} \sum_{i=1}^{N} \cos(\mathbf{E}_{g_i}, \mathbf{E}_{o})
\end{equation}
\vspace{-7mm}

where
\begin{itemize}
    \item $E_{g_i}$ represents the embedding of the generated question $i$.
    \item $E_{o}$ represents the embedding of the original question.
    \item $N$ is the number of generated questions, typically 3 by default.
\end{itemize}

Faithfulness evaluates whether the generated answer accurately represents the information in the retrieved context \cite{ragas-metrics}:
\begin{equation}
    \text{Faithfulness} = \frac{\left| \text{Inferred claims from the given context in generated answer} \right|}{\left| \text{Total claims in the generated answer} \right|}
\end{equation}

%% file: references.bib
@misc{mistralEmbeddingsMistral,
	author = {},
	title = {{E}mbeddings | {M}istral {A}{I} {L}arge {L}anguage {M}odels --- docs.mistral.ai},
	howpublished = {\url{https://docs.mistral.ai/capabilities/embeddings/}},
	year = {},
	note = {[Accessed 11-07-2024]},
}

@misc{ragas-metrics,
	author = {Ragas Documentation},
	title = {Metrics | Component-Wise Evaluation},
	howpublished = {\url{https://docs.ragas.io/en/stable/concepts/metrics/index.html}},
	year = {2024},
	note = {[Accessed July 2024]},
}

@article{Patil2023-fv,
    author={Patil, Rajvardhan and Boit, Sorio and Gudivada, Venkat and Nandigam, Jagadeesh},
    journal={Institute of Electrical and Electronics Engineers Access}, 
    title={A Survey of Text Representation and Embedding Techniques in NLP}, 
    year={2023},
    volume={11},
    pages={36120-36146},
    doi={10.1109/ACCESS.2023.3266377}
}

@misc{openaiTextEmbeddings,
	author = {OpenAI},
	title = {{O}pen {A}{I} {T}ext {E}mbedding {M}odel},
	howpublished = {\url{https://platform.openai.com/docs/guides/embeddings}},
	year = {},
	note = {[Accessed 11-07-2024]},
}

@misc{googleTextEmbeddings,
	author = {},
	title = {{T}ext embeddings {A}{P}{I}, {G}enerative {A}{I} on {V}ertex {A}{I}, {G}oogle {C}loud},
	howpublished = {\url{https://cloud.google.com/vertex-ai/generative-ai/docs/model-reference/text-embeddings-api}},
	year = {},
	note = {[Accessed 11-07-2024]},
}

@misc{promptEng,
	author = {OpenAI},
	title = {{P}rompt engineering},
	howpublished = {\url{https://platform.openai.com/docs/guides/prompt-engineering/strategy-provide-reference-text}},
	year = {},
	note = {[Accessed 12-07-2024]},
}

@article{Han2023-xp,
  author = {Han, Yikun and Liu, Chunjiang and Wang, Pengfei},
  ee = {https://doi.org/10.48550/arXiv.2310.11703},
  journal = {Computing Research Repository},
  title = {A Comprehensive Survey on Vector Database: Storage and Retrieval Technique, Challenge.},
  url = {http://dblp.uni-trier.de/db/journals/corr/corr2310.html#abs-2310-11703},
  volume = {abs/2310.11703},
  year = 2023,
  doi={10.48550/arXiv.2310.11703}
}

@inproceedings{Gao2022-fk,
    title = "Precise Zero-Shot Dense Retrieval without Relevance Labels",
    author = "Gao, Luyu  and
      Ma, Xueguang  and
      Lin, Jimmy  and
      Callan, Jamie",
    editor = "Rogers, Anna  and
      Boyd-Graber, Jordan  and
      Okazaki, Naoaki",
    booktitle = "Proceedings of the 61st Annual Meeting of the Association for Computational Linguistics (Volume 1: Long Papers)",
    month = jul,
    year = "2023",
    address = "Toronto, Canada",
    publisher = "Association for Computational Linguistics",
    url = "https://aclanthology.org/2023.acl-long.99",
    doi = "10.18653/v1/2023.acl-long.99",
    pages = "1762--1777",
}

@inproceedings{Jeong2024-pk,
    title = "Adaptive-{RAG}: Learning to Adapt Retrieval-Augmented Large Language Models through Question Complexity",
    author = "Jeong, Soyeong  and
      Baek, Jinheon  and
      Cho, Sukmin  and
      Hwang, Sung Ju  and
      Park, Jong",
    editor = "Duh, Kevin  and
      Gomez, Helena  and
      Bethard, Steven",
    booktitle = "Proceedings of the 2024 Conference of the North American Chapter of the Association for Computational Linguistics: Human Language Technologies (Volume 1: Long Papers)",
    month = jun,
    year = "2024",
    address = "Mexico City, Mexico",
    publisher = "Association for Computational Linguistics",
    url = "https://aclanthology.org/2024.naacl-long.389",
    doi = "10.18653/v1/2024.naacl-long.389",
    pages = "7036--7050",
}

@misc{zhang2023retrieve,
      title={Retrieve Anything To Augment Large Language Models}, 
      author={Peitian Zhang and Shitao Xiao and Zheng Liu and Zhicheng Dou and Jian-Yun Nie},
      year={2023},
      eprint={2310.07554},
      archivePrefix={arXiv},
      primaryClass={cs.IR},
      url={https://arxiv.org/abs/2310.07554}, 
      doi={10.48550/arXiv.2310.07554}
}

@misc{chen2024bge,
      title={BGE M3-Embedding: Multi-Lingual, Multi-Functionality, Multi-Granularity Text Embeddings Through Self-Knowledge Distillation}, 
      author={Jianlv Chen and Shitao Xiao and Peitian Zhang and Kun Luo and Defu Lian and Zheng Liu},
      year={2024},
      eprint={2402.03216},
      archivePrefix={arXiv},
      primaryClass={cs.CL},
      doi={10.48550/arXiv.2402.03216}
}

@misc{RAG,
      title={Retrieval-Augmented Generation for Large Language Models: A Survey}, 
      author={Yunfan Gao and Yun Xiong and Xinyu Gao and Kangxiang Jia and Jinliu Pan and Yuxi Bi and Yi Dai and Jiawei Sun and Meng Wang and Haofen Wang},
      year={2024},
      eprint={2312.10997},
      archivePrefix={arXiv},
      primaryClass={cs.CL},
      url={https://arxiv.org/abs/2312.10997}, 
      doi={10.48550/arXiv.2312.10997}
}

@misc{wu2024retrievalaugmentedgenerationnaturallanguage,
      title={Retrieval-Augmented Generation for Natural Language Processing: A Survey}, 
      author={Shangyu Wu and Ying Xiong and Yufei Cui and Haolun Wu and Can Chen and Ye Yuan and Lianming Huang and Xue Liu and Tei-Wei Kuo and Nan Guan and Chun Jason Xue},
      year={2024},
      eprint={2407.13193},
      archivePrefix={arXiv},
      primaryClass={cs.CL},
      url={https://arxiv.org/abs/2407.13193},
      doi={10.48550/arXiv.2407.13193}
}

@misc{Lewis,
 author = {Lewis, Patrick and Perez, Ethan and Piktus, Aleksandra and Petroni, Fabio and Karpukhin, Vladimir and Goyal, Naman and K\"{u}ttler, Heinrich and Lewis, Mike and Yih, Wen-tau and Rockt\"{a}schel, Tim and Riedel, Sebastian and Kiela, Douwe},
 booktitle = {Advances in Neural Information Processing Systems},
 editor = {H. Larochelle and M. Ranzato and R. Hadsell and M.F. Balcan and H. Lin},
 pages = {9459--9474},
 publisher = {Curran Associates, Inc.},
 title = {Retrieval-Augmented Generation for Knowledge-Intensive NLP Tasks},
 url = {https://proceedings.neurips.cc/paper_files/paper/2020/file/6b493230205f780e1bc26945df7481e5-Paper.pdf},
 volume = {33},
 year = {2020},
 doi={10.48550/arXiv.2005.11401}
}

@inproceedings{grape,
    title = "Grape: Knowledge Graph Enhanced Passage Reader for Open-domain Question Answering",
    author = "Ju, Mingxuan  and
      Yu, Wenhao  and
      Zhao, Tong  and
      Zhang, Chuxu  and
      Ye, Yanfang",
    editor = "Goldberg, Yoav  and
      Kozareva, Zornitsa  and
      Zhang, Yue",
    booktitle = "Findings of the Association for Computational Linguistics: The 2022 Conference on Empirical Methods in Natural Language Processing",
    month = dec,
    year = "2022",
    address = "Abu Dhabi, United Arab Emirates",
    publisher = "Association for Computational Linguistics",
    url = "https://aclanthology.org/2022.findings-emnlp.13",
    doi = "10.18653/v1/2022.findings-emnlp.13",
    pages = "169--181",
}

@Inproceedings{Amazon,
 author = {Daksha Yadav and Sabrina Zhang and Tom Jin and Prakash Krishnan and Des Clarke},
 title = {Generative AI based virtual assistant for reconciliation research},
 year = {2024},
 url = {https://www.amazon.science/publications/generative-ai-based-virtual-assistant-for-reconciliation-research},
 booktitle = {The Association for the Advancement of Artificial Intelligence 2024 Workshop on AI for Financial Services},
}

@inproceedings{iter,
    title = "Enhancing Retrieval-Augmented Large Language Models with Iterative Retrieval-Generation Synergy",
    author = "Shao, Zhihong  and
      Gong, Yeyun  and
      Shen, Yelong  and
      Huang, Minlie  and
      Duan, Nan  and
      Chen, Weizhu",
    editor = "Bouamor, Houda  and
      Pino, Juan  and
      Bali, Kalika",
    booktitle = "Findings of the Association for Computational Linguistics: The 2023 Conference on Empirical Methods in Natural Language Processing",
    month = dec,
    year = "2023",
    address = "Singapore",
    publisher = "Association for Computational Linguistics",
    url = "https://aclanthology.org/2023.findings-emnlp.620",
    doi = "10.18653/v1/2023.findings-emnlp.620",
    pages = "9248--9274",
}

@article{chen2023benchmarking,
    title={Benchmarking Large Language Models in Retrieval-Augmented Generation},
    volume={38},
    url={https://ojs.aaai.org/index.php/AAAI/article/view/29728},
    doi={10.1609/aaai.v38i16.29728},
    number={16},
    journal={Proceedings of the AAAI Conference on Artificial Intelligence},
    author={Chen, Jiawei and Lin, Hongyu and Han, Xianpei and Sun, Le},
    year={2024},
    month={03},
    pages={17754-17762}
}

@inproceedings{es2023ragas,
    title = "{RAGA}s: Automated Evaluation of Retrieval Augmented Generation",
    author = "Es, Shahul  and
      James, Jithin  and
      Espinosa Anke, Luis  and
      Schockaert, Steven",
    editor = "Aletras, Nikolaos  and
      De Clercq, Orphee",
    booktitle = "Proceedings of the 18th Conference of the European Chapter of the Association for Computational Linguistics: System Demonstrations",
    month = mar,
    year = "2024",
    address = "St. Julians, Malta",
    publisher = "Association for Computational Linguistics",
    url = "https://aclanthology.org/2024.eacl-demo.16",
    pages = "150--158",
}

@conference{csedu20,
    author={Regina Gubareva and Rui Lopes},
    title={Virtual Assistants for Learning: A Systematic Literature Review},
    booktitle={Proceedings of the 12th International Conference on Computer Supported Education - Volume 1: CSEDU,},
    year={2020},
    pages={97-103},
    publisher={SciTePress},
    organization={Institute for Systems and Technologies of Information, Control and Communication},
    doi={10.5220/0009417600970103},
    isbn={978-989-758-417-6},
}

@article{llm_long_context,
  publtype={informal},
  author={Tianle Li and Ge Zhang and Quy Duc Do and Xiang Yue and Wenhu Chen},
  title={Long-context LLMs Struggle with Long In-context Learning},
  year={2024},
  cdate={1704067200000},
  journal={Computing Research Repository},
  volume={abs/2404.02060},
  url={https://doi.org/10.48550/arXiv.2404.02060},
  doi={10.48550/arXiv.2404.02060}
}

@article{Farquhar2024,
    author={Farquhar, Sebastian
    and Kossen, Jannik
    and Kuhn, Lorenz
    and Gal, Yarin},
    title={Detecting hallucinations in large language models using semantic entropy},
    journal={Nature},
    year={2024},
    month={Jun},
    day={01},
    volume={630},
    number={8017},
    pages={625-630},
    issn={1476-4687},
    doi={10.1038/s41586-024-07421-0},
    url={https://doi.org/10.1038/s41586-024-07421-0}
}

@inproceedings{RAG_survey_july2024,
    author = {Fan, Wenqi and Ding, Yujuan and Ning, Liangbo and Wang, Shijie and Li, Hengyun and Yin, Dawei and Chua, Tat-Seng and Li, Qing},
    title = {A Survey on RAG Meeting LLMs: Towards Retrieval-Augmented Large Language Models},
    year = {2024},
    isbn = {9798400704901},
    publisher = {Association for Computing Machinery},
    address = {New York, NY, USA},
    url = {https://doi.org/10.1145/3637528.3671470},
    doi = {10.1145/3637528.3671470},
    booktitle = {Proceedings of the 30th ACM SIGKDD Conference on Knowledge Discovery and Data Mining},
    pages = {6491–6501},
    numpages = {11},
    location = {Barcelona, Spain},
    series = {Knowledge Discovery and Data Mining '24}
}
